\documentclass{article}
\title{Generalization Limits of In-Context Operator Networks \\ for Higher-Order Partial Differential Equations}
\author{
    Jamie Mahowald\footnote{Department of Mathematics, University of Texas at Austin, Austin, Texas (j.mahowald@utexas.edu).}
}
\date{}
\usepackage{multicol}
\usepackage[letterpaper, margin=1in]{geometry}
\usepackage[hidelinks]{hyperref}
\usepackage{caption}
\usepackage{subcaption}
\usepackage[T1]{fontenc}
\usepackage{multicol}
\usepackage{appendix}
\usepackage{amsmath, amsfonts, amssymb, amsthm}
\usepackage{graphicx}
\usepackage{pdflscape}

\begin{document}
    \maketitle
    \vspace{-3em}
    \begin{center}
        \textit{Project Adviser:} Tan Bui-Thanh\footnote{Department of Aerospace Engineering and Engineering Mechanics, Oden Institute for Computational Engineering and Sciences, University of Texas at Austin, Texas (tanbui@oden.utexas.edu, https://users.oden.utexas.edu/tanbui/).}
    \end{center}

\begin{abstract}
We investigate the generalization capabilities of In-Context Operator Networks (ICONs), a new class of operator networks that build on the principles of in-context learning, for higher-order  partial differential equations. We extend previous work by expanding the type and scope of differential equations handled by the foundation model. We demonstrate that while processing complex inputs requires some new computational methods, the underlying machine learning techniques are largely consistent with simpler cases. Our implementation shows that although point-wise accuracy degrades for higher-order problems like the heat equation, the model retains qualitative accuracy in capturing solution dynamics and overall behavior. This demonstrates the model's ability to extrapolate fundamental solution characteristics to problems outside its training regime.
\end{abstract}

\tableofcontents

\section{Introduction}
Using neural networks to solve differential equations is a natural evolution in the long development of numerical methods and in the more recent advances in machine learning. These methods are particularly useful for equations that have no analytical solutions or that are computationally prohibitive to derive analytically. Furthermore, the structure of numerical data makes these equations and their solutions intuitive inputs to neural networks.

Several model architectures have leveraged different aspects of this data type to produce accurate and generalizable solutions: physics-informed machine learning \cite{piml_2021}, for instance, incorporates information about the governing equations into the model loss, while operator networks like DeepONet \cite{deeponet_2021} use the operator structure of differential equations to restructure the input-output sequence. But models built under both these frameworks must be retrained to handle new equations or operators.

\textit{In-context operator networks} (ICONs) \cite{icon_1_original} apply the in-context learning technique to avoid the retraining limitation. By demonstrating several examples of the same operator at inference time before requesting a solution, the model can effectively learn the operator and apply it to the question, expanding the generalization ability of these models without sacrificing accuracy. 

In this work, we extend the type and scope of ordinary and partial differential equations (ODEs and PDEs) provided to the foundation ICON model using numerical methods and examine their performance on higher-order and higher-dimensional differential equations. We show that processing these more complex inputs does not require a change in the underlying machine learning techniques from previous inputs. But owing to the complexity of these problems, new computational methods are needed to train on them efficiently. We also examine the difference between model inference on in-distribution and out-of-distribution problems to quantify the limits of generalization for these models. In-distribution problems are those that the model has seen during training, while out-of-distribution problems are those that differ significantly from the training data. Lastly, we introduce a user interface for equation solutions using a pre-trained model and show that, while model-evaluated solutions are sufficient for simple problems, they do not yet outperform their numerical counterparts for equations with finer meshes. 

The rest of this paper is organized as follows: Section 2 reviews the foundational work in operator learning and introduces our target equation class. Section 3 details our methodology for data generation and model training. Section 4 presents our empirical results and analysis. Section 5 discusses the implications of our findings and potential future directions. Finally, Section 6 concludes with a summary of our key insights.
\section{In-context operator learning for differential equation}

In-context learning is a machine learning technique that allows a model to see a set of task demonstrations -- a context -- from which it can learn a task and apply it to a question without explicit retraining. Operator learning is any technique that allows models to learn operators that map between function spaces. The combination of these -- in-context operator learning -- is then a framework that allows models to see a set of equations characterized by a single operator, from which it can learn that operator and apply it to the equation being solved for, known as the question equation. For instance, in a forward time-dependent PDE, the question equation may be an initial value, and the model output will be a solution at a later timestep based on the operator defined by the PDE.

\subsection{In-context operator networks}
\subsubsection{In-context operator learning with data prompts for differential equation problems}

The initial development of the In-Context Operator Network (ICON) framework, first introduced by Liu et al. in \cite{icon_1_original}, established a foundational approach to operator learning through a contextual key-value representation system. The architecture implements an encoder-decoder neural network structure inspired by GPT-3, demonstrating capabilities across multiple operational domains:

\begin{itemize}
    \item \textbf{In-distribution operator performance}: The framework exhibits strong generalization capabilities within the training distribution, maintaining low relative errors across diverse problem sets.
    \item \textbf{Resolution adaptability}: Demonstrated flexibility in handling varying densities of key-value pairs during inference, encompassing both super-resolution and sub-resolution scenarios.
    \item \textbf{Out-of-distribution generalization}: Measurable performance when processing operators outside the training distribution, maintaining some effectiveness across parameter variations, although largely not across problem types.
    \item \textbf{Novel equation handling}: Some capacity to predict on unseen equation forms.
\end{itemize}

\subsubsection{Language model integration for multi-modal differential equation solving}

Later developments introduced ICON-LM \cite{icon_2_finetune}, an evolution of the original framework incorporating natural language processing capabilities through GPT-2 fine-tuning. This advancement introduced "captions" as supplementary input features, enabling natural language descriptions of PDE problems alongside traditional numerical inputs.

The architecture underwent significant optimization, consolidating the previous encoder-decoder structure into an encoder-only transformer. Training follows a next-token prediction method characteristic of language models, implementing a specialized transformer mask to prevent unintended attention-based information leakage.

This model improved significantly over the original ICON framework, particularly in scenarios using "precise captions" - natural language inputs with explicit parameter specifications. The enhancement also improved zero-shot and few-shot learning capabilities.

\subsubsection{Conservation law applications in PDE contexts}

Further research \cite{icon_3_pdegen} explored the framework's generalization capabilities to 1D scalar nonlinear conservation laws of the form
\[\partial_t u(t,x) + \partial_x f(u(t,x)) = 0, x \in [0,1],\]
with periodic boundary conditions. The forward operator \(\mathcal{F}_{f, \tau}\) and reverse operator \(\mathcal{R}_{f, \tau}\) are defined as:

\begin{align*}
  &\mathcal{F}_{f, \tau} [u(0, \cdot)] = u(\tau, \cdot) \quad \text{ s.t. } \quad \partial_t u(t,x) + \partial_x f(u(t,x)) = 0, \\
  &\mathcal{R}_{f, \tau} [u] = \{v \mid \mathcal{F}_{f, \tau} [v] = u\}
\end{align*}

The implementation used a cubic flux function for training:
\[f(u) = au^3 + bu^2 + cu, \implies \partial_t u + \partial_x (au^3 + bu^2 + cu) = 0.\]

Performance evaluation encompassed four key aspects:

\begin{itemize}
    \item \textbf{In-distribution Performance}: Error rates decreased with increasing example counts (1-5) for operators evaluated at future timestep \(\tau=0.1\).
    \item \textbf{Novel PDE Adaptation}: The framework demonstrated effective generalization to alternative flux functions (e.g., \(f(x) = sin(u) - cos(u)\)), with error accumulation over temporal progression.
    \item \textbf{Variable Transformation}: Performance under affine transformation \(v = \alpha u + \beta\) showed improved results with parameters approaching the training distribution.
    \item \textbf{Stride Variation}: Analysis revealed inverse correlation between error rates and stride magnitude.
\end{itemize}

\subsection{Problem setup}

Like older operator networks, this approach treats differential equation solving as an operator learning problem. We decompose each differential equation listed in \ref{appendix:ode-pde-forms} into three components: \textbf{parameters} (known constants, distinct from internal model parameters), \textbf{conditions} (known functions or initial/boundary values), and \textbf{quantities of interest} (QoIs, what we want to solve for). A neural network then learns the operator that maps conditions to QoI for given parameters. To some degree this division is up to the user, but in general, for a forward problem, parameters are constants, conditions are control functions, and QoIs are solutions. For an inverse problem, the roles of conditions and QoIs are switched.

We formalize this as an operator \(\mathcal{F}\) that depends on parameters \((a_1, a_2)\) and maps condition functions \(c(t)\) to solutions \(u(t)\): \(\mathcal{F}_{(a_1, a_2)}[c(t)] = u(t)\). The inverse problem concerns \(\mathcal{R}\), the inverse of \(\mathcal{F}\), where \(\mathcal{R}_{(a_1, a_2)}[u(t)] = c(t)\) if \(c(t)\) is unique. If not, \(\mathcal{R}_{(a_1,a_2)}[u(t)] = \{c(t): \mathcal{F}_{(a_1, a_2)}[c(t)] = u(t)\}\).
If we fix \((a_1, a_2)\) (i.e., define an operator) and generate \(n\) control functions \(c_i(t)\) corresponding to \(n\) solutions \(u_i(t)\), \(1 \leq i \leq n\), then for the operator \(\mathcal{F}_{(a_1, a_2)}\), the operator solves the system:
\[
\begin{aligned}
  &\frac{d}{dt} u_1(t) = a_1 c_1(t) + a_2 
    &&\qquad \Longleftrightarrow\qquad 
    \mathcal{F}_{(a_1, a_2)}[c_1(t)] = u_1(t) \\
  &\frac{d}{dt} u_2(t) = a_1 c_2(t) + a_2 
    &&\qquad \Longleftrightarrow\qquad 
    \mathcal{F}_{(a_1, a_2)}[c_2(t)] = u_2(t) \\
  &\vdots 
    &&\qquad \vdots \\
  &\frac{d}{dt} u_n(t) = a_1 c_n(t) + a_2 
    &&\qquad \Longleftrightarrow\qquad 
    \mathcal{F}_{(a_1, a_2)}[c_n(t)] = u_n(t)
\end{aligned}
\]

Extending to partial differential equations: a more traditional PDE governed by initial and boundary conditions, like the heat equation in a one-dimensional rod with a source term proportional to the solution, is given by
\[
\begin{cases}
    \text{PDE: } & \frac{\partial u}{\partial t} + k\frac{\partial^2 u}{\partial x^2} + \alpha u = 0, \\
    \text{Boundary conditions: } & u(0,t)=u(L,t)=0, \\
    \text{Initial condition: } & u(x,0)=f(x).
\end{cases}
\]
Lacking a defined control function \(c(t)\), we may define the condition as the initial condition \(f(x)\), while the parameters defining the operator are constants \(k, \alpha\) and boundary conditions \(u(0,t), u(L,t)\). Constant parameters are not strictly required, but spatial parameters (e.g., \(k(x), \alpha(x)\)) significantly increase memory complexity. We list all our equation types, with their associated neural operators, in \ref{appendix:ode-pde-forms}

\subsubsection{Neural vs. differential operators}
Using the example of the heat equation, it is important to distinguish the differential operator 
\[L := \frac{\partial}{\partial t} + k\frac{\partial^2}{\partial x^2} + \alpha \mathbf{I},\]
parameterized by \(k, \alpha\) and satisfying \(L[u]=0\), from the neural operator \(\mathcal{F}[y]\) parameterized by the same \(k, \alpha\) and \(u(0,t), u(L,t)\) and satisfying \[\mathcal{F}[u(x,0)] = u(x,t).\]

For example, for the self-adjoint operator formulation of the Sturm-Liouville equation,
\[\frac{d}{dx} \left[p(x)\frac{dy}{dx}\right] + q(x)y + \lambda w(x)y = 0\]
where the linear self-adjoint operator
\[L = \frac{d}{dx} \left[ p(x) \frac{d}{dx}\right] + q(x)\]
satisfies \[L[y] + \lambda w(x)y = 0,\]
we may conceivably define any combination of the several quantities as \(p(x), q(x), w(x)\) as parameters and reserve at least one for a condition, along with the boundary values \(y(a)\) and \(y(b)\), depending on the unknowns of the problem. For a simple operator where \( p(x)=p_0 \), \( q(x)=q_0 \), and \( w(x)=w_0 \) are constants, natural choice is to define the parameters \( (\lambda, p_0, q_0, w_0) \), then model the operator \(\mathcal{F}\) that maps boundary conditions to the eigenfunction \( y(x) \) across the domain. This decomposition is independent of the differential operator's natural structure and is chosen based on the problem requirements.
\section{Data generation and model architecture}

To train our operator learning framework across the 19 equation types, we require large datasets of (condition, QoI) pairs for various parameter values. Since analytical solutions are rarely available for more involved differential equations, we generate synthetic training data using established numerical methods, ensuring our neural networks learn from mathematically consistent operator mappings.

\subsection{Synthetic data generation using numerical methods}

\subsubsection{Data-generation using traditional numerical solvers}
Most of the ground-truth data for the 19 differential equation types used in experiments in \cite{icon_1_original} are generated first by randomly sampling the parameters and conditions, and then by solving for \(u\) using traditional numerical methods. A condition in the form of a control function is most often denoted \(c(t) \text{ or } c(x).\) Take a simple linear ODE: \(\frac{d}{dt} u(t) = a_1 c(t) + a_2\), defined on \([0,L]\) for \( L>0 \). A single operator for this equation is defined by randomly generated \(a_1\) and \(a_2\). For each of these operators, we generate multiple condition-QoI pairs using a two-dimensional Gaussian process with a high-variance radial basis function (RBF) kernel. The corresponding solution \(u(t)\) is then calculated numerically using the Euler method. Boundary conditions \(u(0) \text{ and } u(L)\) and control function \(c(t)\) are then recorded together as the condition for that demo, and \(u(t)\) is recorded as the QoI. 

Poisson problems and linear reaction-diffusion problems are handled similarly, except that the standard finite-difference methods like the tridiagonal matrix algorithm are used to solve for \(u(x)\) instead of the Euler method. Thus, data for these problems is handled analogously to how one would solve them in practice: given a condition, some known constants, and a differential equation, find a solution that satisfies that equation.

Data for the conservation law 
\[\partial_t u(t,x) + \partial_x f(u(t,x)) = 0, x \in [0,1]\]
is generated with the weighted essentially non-oscillatory method \cite{weno}, a high-resolution numerical scheme designed for hyperbolic PDEs with sharp gradients.

\subsubsection{Data-generation by starting with the solution \(u\)}

In contrast, the data for damped oscillator and the nonlinear reaction-diffusion problems are generated by \textit{starting} with the solution \(u(x)\), then using finite-difference approximation to calculate the derivatives, and finally substituting these into the rest of the equation. Traditional forward solvers for these complex equations often suffer from numerical artifacts and stability issues on the desired domains. Our goal is to generate high-quality synthetic datasets rather than simulating specific physical scenarios directly, so we can exploit this freedom by starting with mathematically well-behaved solutions and working backwards to determine the corresponding input conditions. 

For nonlinear R-D, \(-\lambda a \frac{d^2}{dx^2} u(x) + ku(x) = c(x)\) on \([0,L]\), each operator is defined by boundary conditions \(u(0), u(L)\) and constants \(k, a\) (\(\lambda\) is set to 0.05 for all problems). For each demo, \(u(x)\) is generated by Gaussian process and modified to fit the boundary conditions. Next, \(\frac{d^2}{dx^2} u(x)\) is calculated using finite differences. Then, \(c(x)\) is calculated simply by substituting \(u(x)\) and its numerical derivatives into the equation. Finally, \(c(x)\) is recorded as the condition and \(u(x)\) is recorded as the QoI. 

For higher-order, higher-dimensional partial differential equations, we use the simple (though not physical) formulation 
\[
au_{xx}(x,t) + bu_{xt}(x,t) + cu_{tt}(x,t)
+ du_x(x,t) + eu_t(x,t) + fu(x,t) = g(x,t),
\]
where the parameters \(a, ..., f\) are bounded real numbers.

The objective for data generation is to run the same \(u\)-first method used for damped oscillator and nonlinear R-D problems. We generate initial conditions \(u(x,0), u(x,1)\) by 2D Gaussian process, and \(u(x,t)\) by 3D Gaussian process, interpolating \(u(x,t)\) to the initial conditions by
\[v(x,t) := u(x,t) + (1-t)[v(x,0)-u(x,0)] + t[v(x,1)-u(x,1)].\] We approximate \(u_{xx}, u_{xt}, u_{tt}, u_t, u_x\) via finite difference and calculate \(g(x,t)\) directly. We record \(g(x,t)\) as the condition and \(u(x,t)\) as the QoI (see Fig. \ref{fig:3dexample}). The neural network's objective is then to determine the operator \(\mathcal{F}\) defined by \((a,...,f)\) that maps \(g(x,t)\) to \(u(x,t)\).\footnote{The full repository fork can be found at \url{https://github.com/j-mahowald/in-context-operator-networks}.}

\label{fig:3dexample}
\begin{figure}[!htbp]
\begin{center}
\includegraphics[width = 0.8\linewidth]{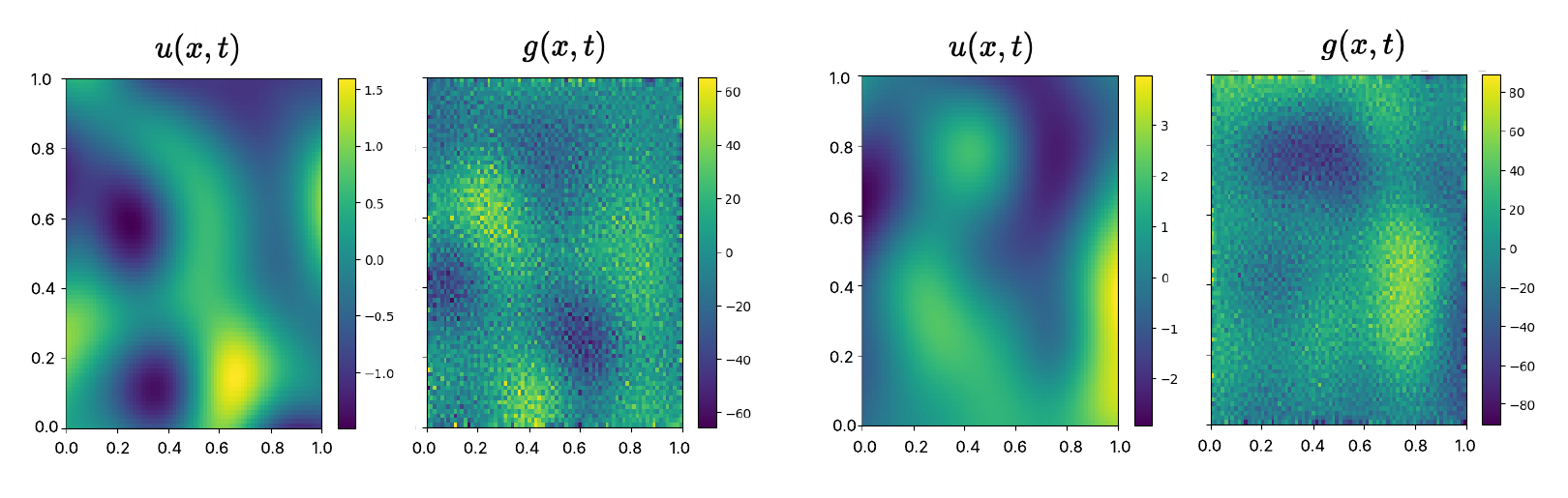}
\end{center}
\caption{Two samples from a single operator defined by \((a, ..., f)\). In this case \((a,b,c,d,e,f) = (0.4563, 0.1500, -0.4341, -0.0525, -0.0457, 0.1578)\). These were generated from a Gaussian process with an RBF kernel with length scale 0.2 (one fifth of the domain) and variance 2.0.}
\end{figure}

\subsection{Model architecture}
The implemented transformer architecture consists of 6 layers with 8 attention heads. Each head maintains a dimension of 256, matching the model's primary dimension. The architecture employs a widening factor of 4, resulting in a hidden dimension of 1024. The model utilizes vanilla attention mechanisms and Glorot uniform kernel initialization, with no dropout implemented (rate = 0). 

The model underwent training for 1,000,000 steps, distributed across 100 epochs (10,000 steps per epoch), requiring approximately 50 hours of computation time (see \ref{appendix:training-details} for details). Training and testing losses were monitored throughout the process to assess model convergence and generalization capabilities (see Fig. \ref{fig:train-test}). 
\section{Results}
\subsection{Results and analysis}
\begin{figure}[!htbp]
\begin{center}
\includegraphics[width=0.40\linewidth]{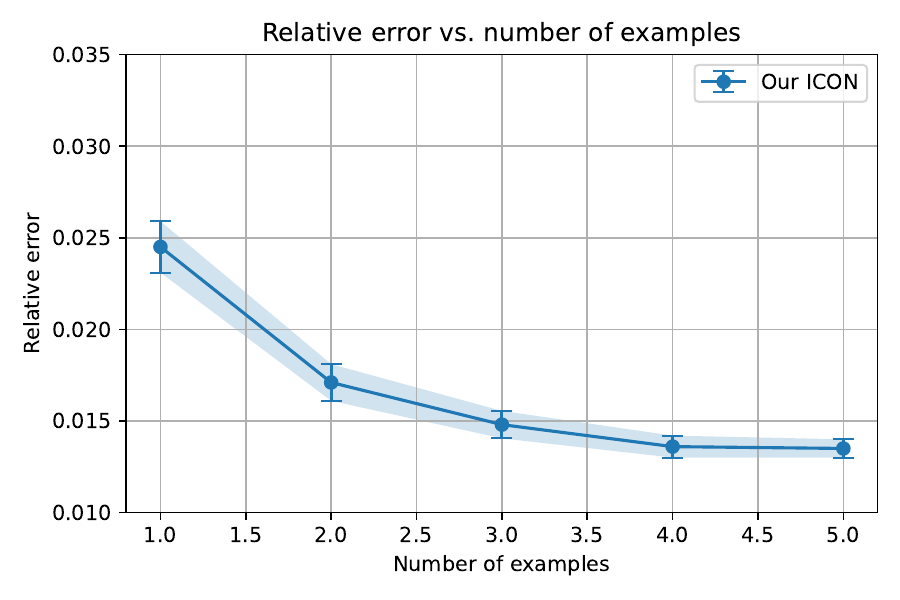}
\includegraphics[width=0.40\linewidth]{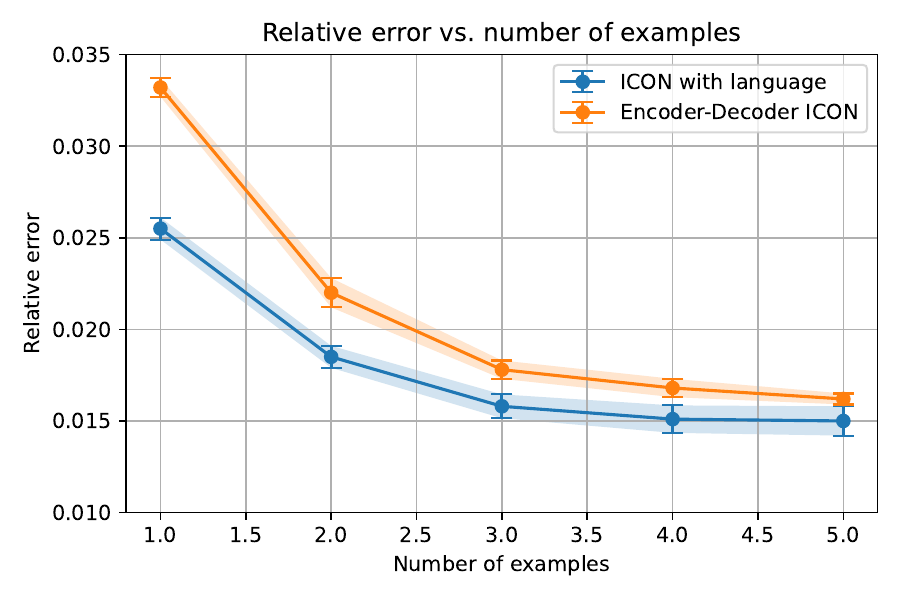}
\end{center}
\caption{We achieve similar average error as the original paper on inference across context sizes. Left: average across all 19 problems of the average testing error for each problem (i.e., the average of the averages), with similarly defined error bars. This can be compared to the original experiment's error \cite{icon_1_original} on the right.}\label{fig:error}
\end{figure}

Our model's performance is comparable to or better than that of the original implementation across 19 test problems. The average testing error, computed as the mean across all problem-specific averages, closely aligns with the benchmark results from \cite{icon_1_original} (see Fig. \ref{fig:error}).

\begin{figure}[!htbp]
\begin{center}
\includegraphics[width=0.85\linewidth]{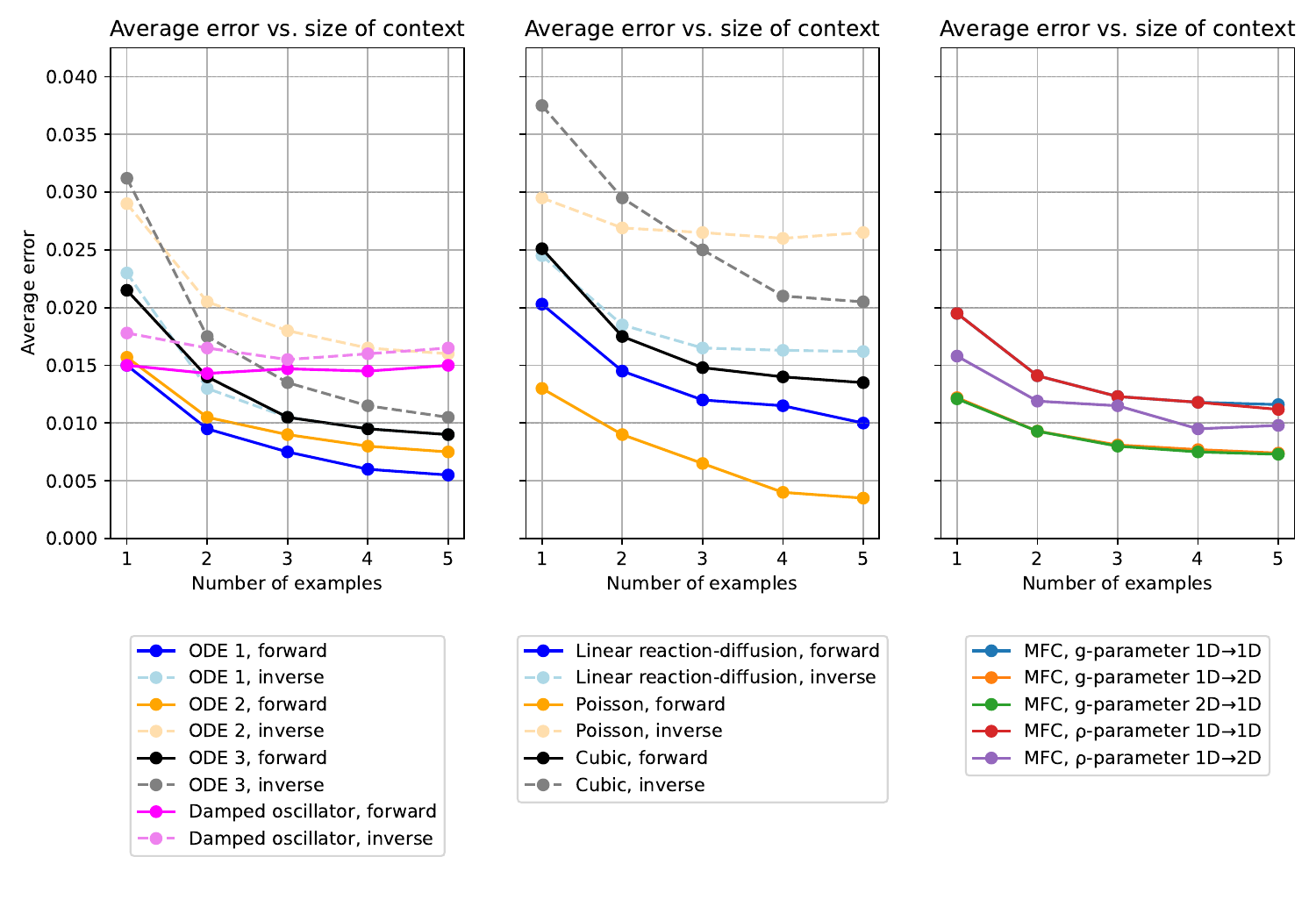}
\caption{Average error in inference vs. the number of samples provided, split into three panels for visibility. See \ref{appendix:ode-pde-forms} for full forms. We see that forward problems tend to achieve higher accuracy than their inverse counterparts. In the ICON setting, accuracy is uncorrelated with the complexity of the relationship: intricate MFC problems achieve similar or lower error as simpler ODEs, while intermediate PDEs have a wide range.}
\label{fig:all-error}
\end{center}
\end{figure}

Dividing into particular equation types reveals a similar downward trend. However, certain problem types, including the damped oscillator and Poisson equations, did not improve in accuracy as samples increased, due to the heterogeneity of in-context examples for these problem types (see Fig. \ref{fig:all-error}). In particular, a damped oscillator has more step-by-step variability than a monotonic solution to a simple ODE. This suggests that the model may struggle to generalize across highly heterogeneous examples within a problem type.

\subsection{Model evaluation}
\begin{figure}[!htbp]
\begin{center}
    \includegraphics[width=0.9\linewidth]{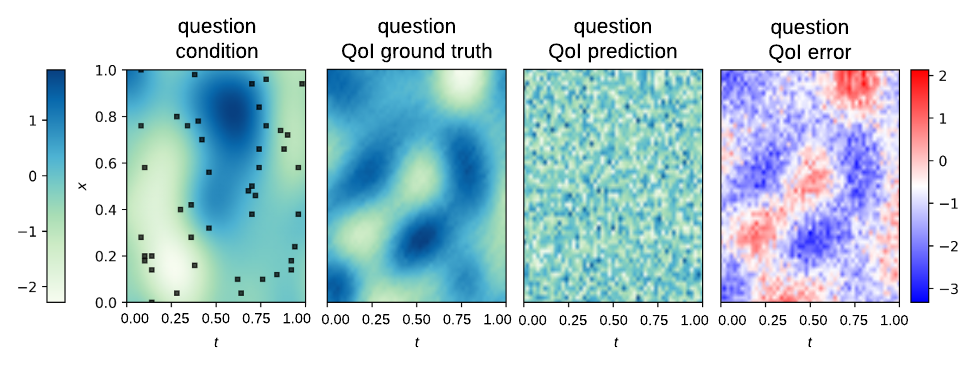}
    \includegraphics[width=0.4\linewidth]{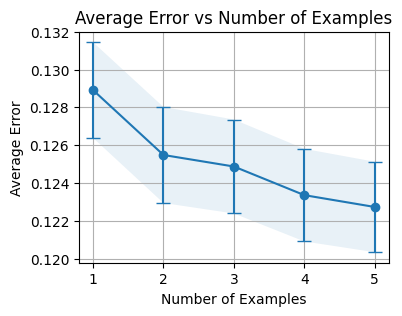}
    \includegraphics[width=0.475\linewidth]{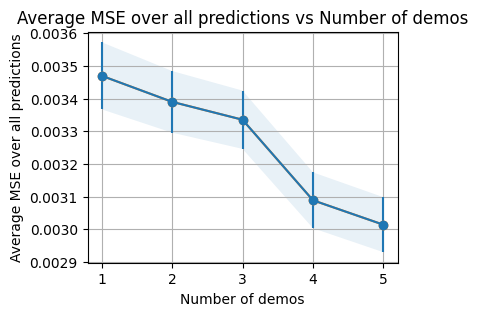}
\end{center}
\caption{The model is kept from making entirely random predictions by the transformer's accurate detection of global patterns, sometimes at the expense of local shapes. The model input is shown on the left, where individual input points are randomly selected (black squares). The model is expected to infer on future points based on this context. While the model fails to discern individual features (third panel), its \textit{average} predictions follow those of the ground truths they aim to predict. The mean-squared difference between the averages of the prediction and the ground truth (bottom right) is much lower than the error in individual tokens (bottom left), computed as a Frobenius norm of the element-wise difference between the prediction and ground-truth matrices.}\label{fig:global-pattern}
\end{figure}

The model's performance on out-of-distribution tasks, particularly the 3D linear PDE problem, reveals limitations in generalization. While prediction errors for these cases are an order of magnitude higher than in-distribution problems, the model is capable of capturing global patterns (see Fig. \ref{fig:global-pattern}), though not local features. This is evidenced by the lower mean-squared difference between averaged predictions and ground truths compared to individual token errors.

\begin{figure}
\begin{center}
    \begin{subfigure}[b]{0.4\textwidth}
        \includegraphics[width=\linewidth]{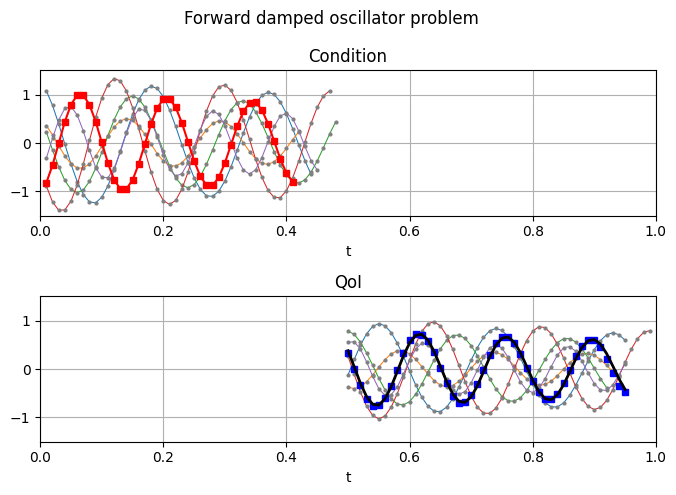}
        \caption*{(a)}
        \label{fig:b}
    \end{subfigure}
    \begin{subfigure}[b]{0.4\textwidth}
        \includegraphics[width=\linewidth]{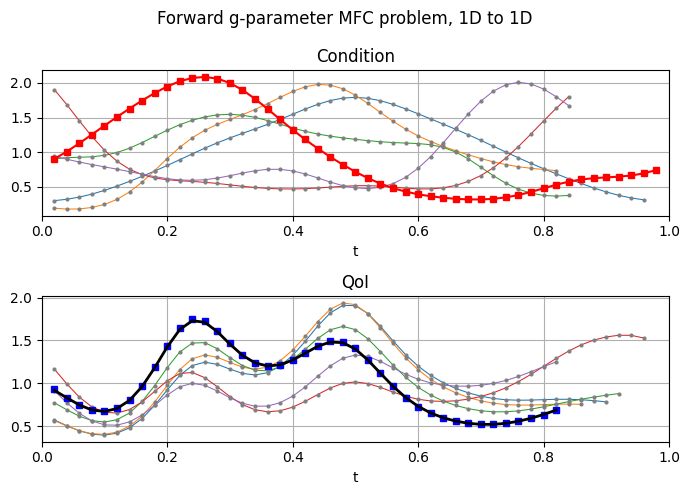}
        \caption*{(b)}
        \label{fig:c}
    \end{subfigure}
    
    \begin{subfigure}[b]{0.4\textwidth}
        \includegraphics[width=\linewidth]{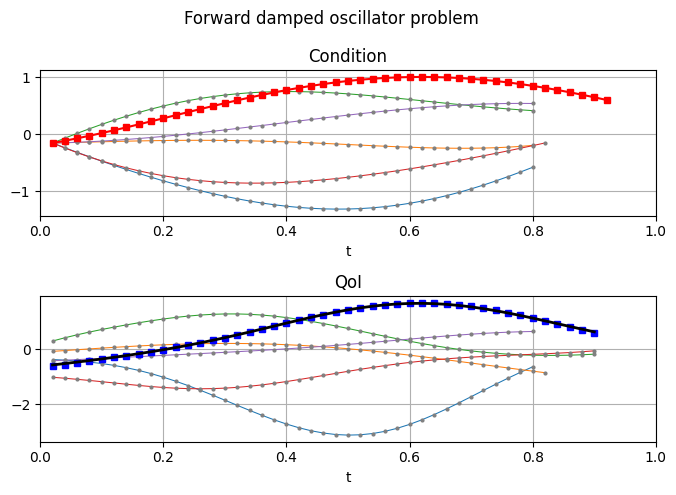}
        \caption*{(c)}
        \label{fig:d}
    \end{subfigure}
    \begin{subfigure}[b]{0.4\textwidth}
        \includegraphics[width=\linewidth]{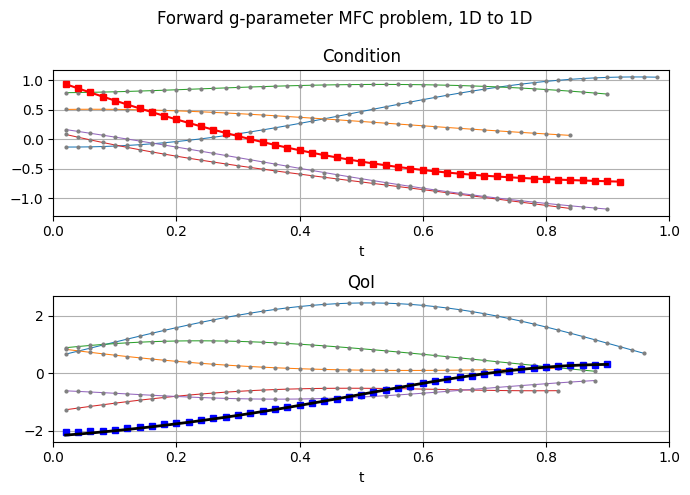}
        \caption*{(d)}
        \label{fig:e}
    \end{subfigure}
\end{center}
\caption{Inference on several problems: (a) forward damped-oscillator, where the model is supplied with example conditions and quantities of interest (light colors) and the question condition (bright red) and is expected to predict the question quantity of interest (dark blue), with ground-truth QoI shown in black; (b) forward mean-field control problem; (c) forward ordinary differential equation; (d) backward ordinary differential equation.}
\label{fig:examples}
\end{figure}

Nevertheless, accuracy remains high for in-distribution inputs. In particular, for temporally extrapolated inferences (i.e., the user inputs data for some length of time, and the model is expected to continue the data for another length of time), the model is capable of inferring across a small domain gap, if supplied with sufficient context (see Fig. \ref{fig:examples} (a)).

\pagebreak
\subsection{Out-of-distribution case study: heat equation}

We run inference on several instances of the heat equation \(u_t=k \cdot u_{xx}+\alpha u\), where \(a=0\) for the homogeneous case. Translated into ICON terms, the condition is set as the initial condition \(u(x,0)\), the quantity of interest as \(u(x, \tau)\) for some predetermined time step \(\tau\), and the parameters are the diffusivity constant \(k\) and time-independent boundary conditions \(u(0)\) and \(u(L)\) for domain length \(L\). The operator \(\mathcal{F}\) then satisfies \(\mathcal{F}_{k, u_0, u_L}[u(x,0)] = u(x, \tau)\).

We see that even a model that is \textit{not} trained on this equation is still capable of directionally accurate inference.

\begin{figure}[!htbp]
\begin{center}
\includegraphics[width=0.90\linewidth]{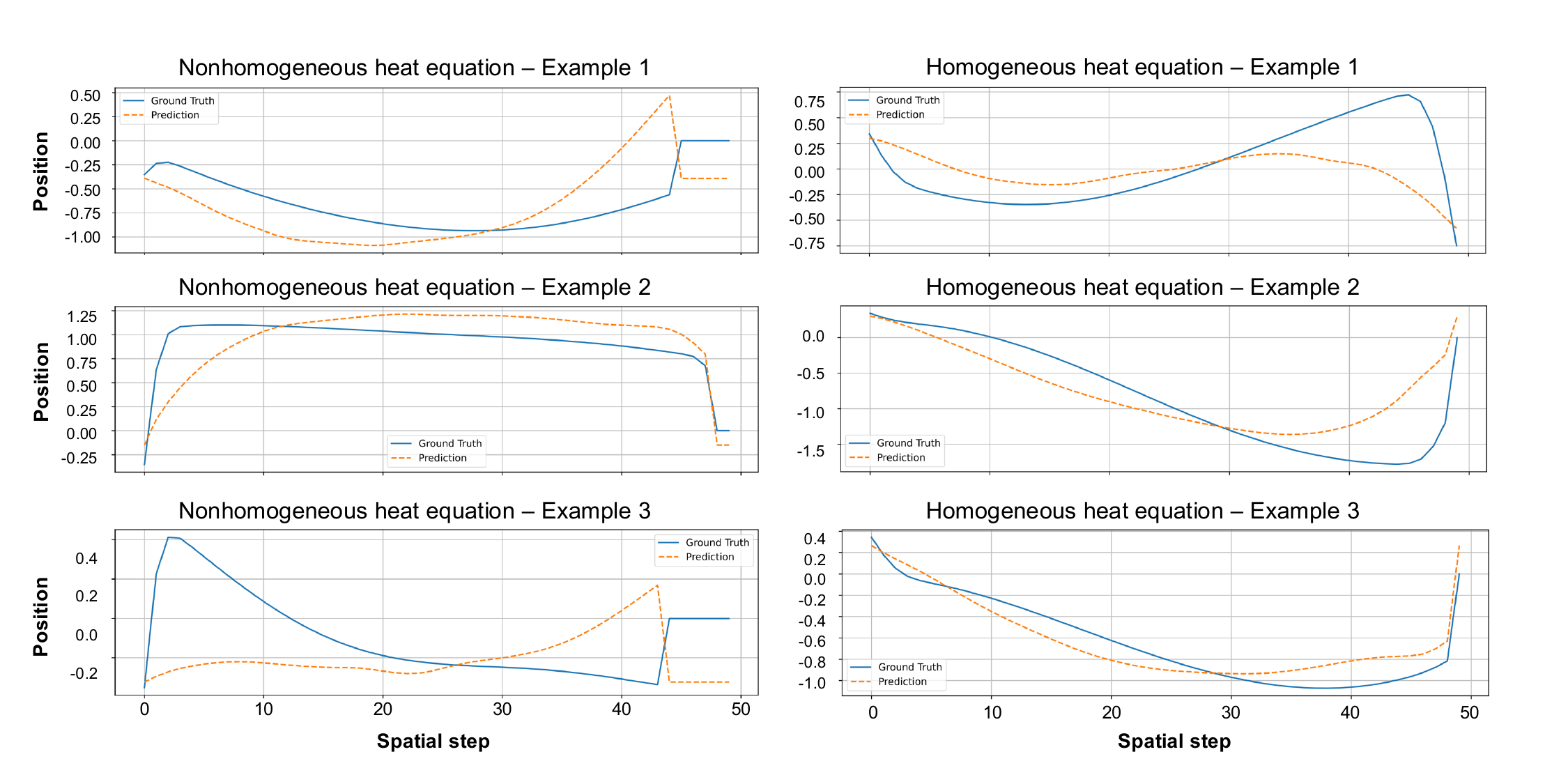}
\end{center}
\caption{Inference on homogeneous (left) and nonhomogeneous (right) heat equations \(u_t=k \cdot u_{xx}+\alpha u\) for timestep \(\tau\)=0.1 is directionally correct in 5 of 6 randomly selected examples. Boundary conditions \(u_0, u_L\) are uniformly chosen in \([-1.0, 1.0]\), \(k\) in \([0.001, 0.01]\), and (in the nonhomogeneous case) \(\alpha\) in \([-0.01, -0.001]\) for stability.}
\label{fig:heat-inference}
\end{figure}

Furthermore, compared to in-distribution inference, out-of-distribution inference is much less dependent on the number of examples provided in the context. 

\begin{figure}[!htbp]
\begin{center}
    \includegraphics[width=0.40\linewidth]{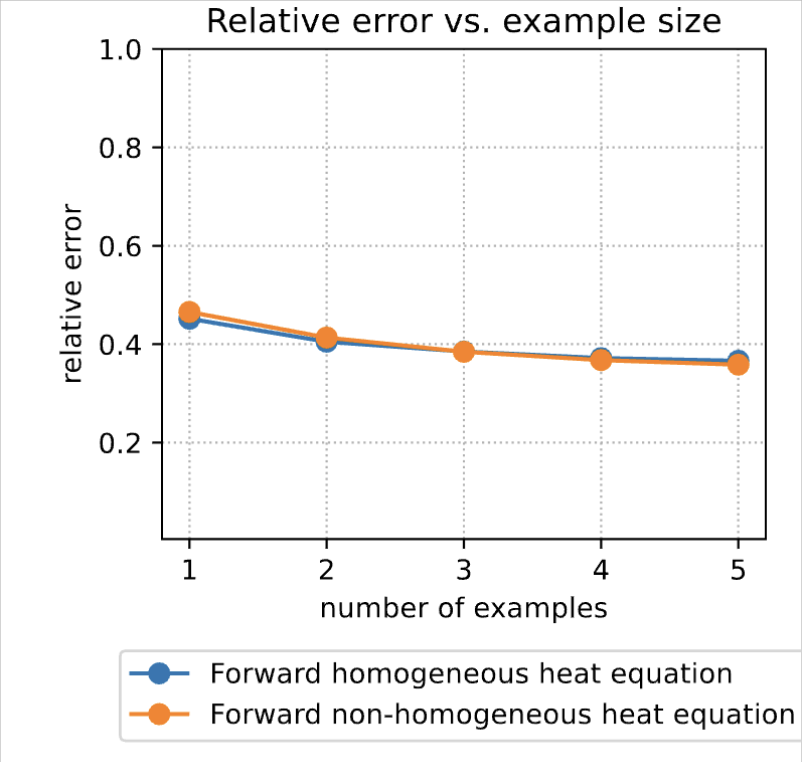}
    \caption{Relative error on the forward homogeneous and nonhomogeneous heat equation.}
\end{center}
\end{figure}

This suggests that the model's performance on out-of-distribution tasks relies more heavily on its pretrained understanding of differential equations rather than the in-context examples provided. While in-distribution tasks show clear improvement with additional examples (see again Fig. \ref{fig:error}), the heat equation results maintain relatively constant error rates regardless of context size. 

The predictions show a tendency to smooth out sharp transitions more aggressively than the ground truth solutions, with the model struggling to maintain the correct boundary conditions at inflection points. This systematic error pattern suggests that while the model has learned general principles of diffusive processes, it may be biased towards solutions with certain smoothness properties that don't fully align with the true physics of the heat equation.

Most notably, the model's predictions consistently underestimate the magnitude of solution variations, producing more conservative estimates that tend toward the mean value. This behavior is particularly evident in Example 1, where the prediction fails to capture the full amplitude of both positive and negative excursions in the solution. These limitations point to potential areas for improvement in the model's ability to handle out-of-distribution boundary value problems.
\section{Discussion}
\subsection{Interactive implementation}
An interactive framework has been developed to facilitate the practical application of the model. The implementation accepts differential equation specifications via a structured dictionary format, which includes equation type, domain discretization, parameters, and boundary conditions. The system autonomously generates demonstration cases to support operator learning before producing predictions using the pre-trained model.

For example, the framework can process linear ODEs of the form:
\[
u'(t) = 0.5u(t) + 0.25c(t) - 0.3, \quad u(0) = 0, \quad c(t) = \frac{1}{2}x
\]

The implementation currently operates locally, with a demonstration available via external repository\footnote{A demonstration of the implementation can be accessed at: \url{https://drive.google.com/file/d/1R99NVhD2S0bZosK5eyu5kR24dZsT8aya/view?usp=sharing}}. This framework represents a step toward making the model accessible for practical differential equation-solving applications.

\subsection{Comparison with traditional solvers}

\begin{figure}[!htpb]
\begin{center}
    \begin{subfigure}[b]{\textwidth}
        \centering
        \includegraphics[width=0.8\linewidth]{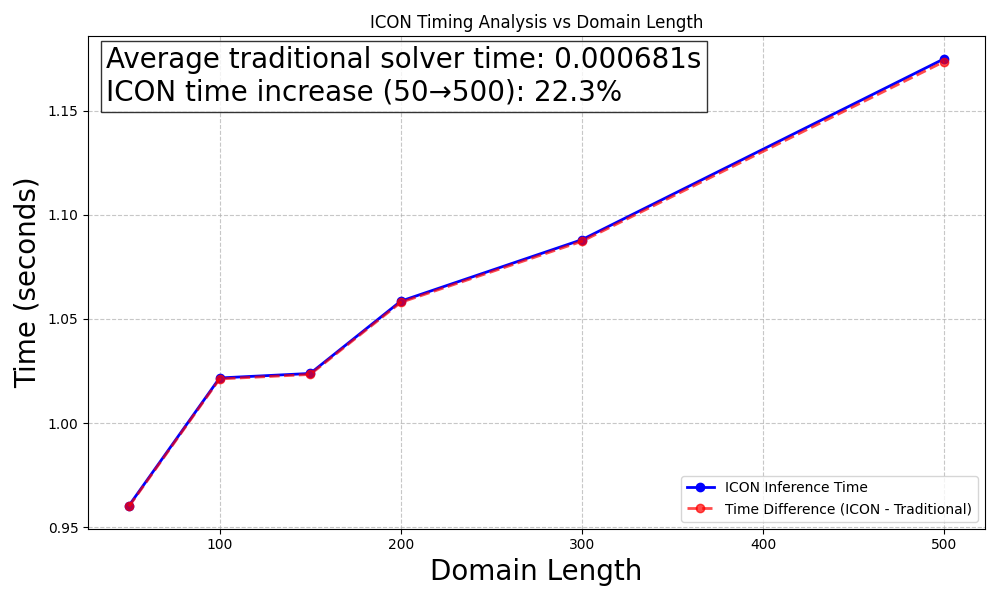}
        \caption{Time taken for inference on a Poisson problem compared with the length of the domain.}
    \end{subfigure}
    
    \begin{subfigure}[b]{\textwidth}
        \centering
        \includegraphics[width=0.75\linewidth]{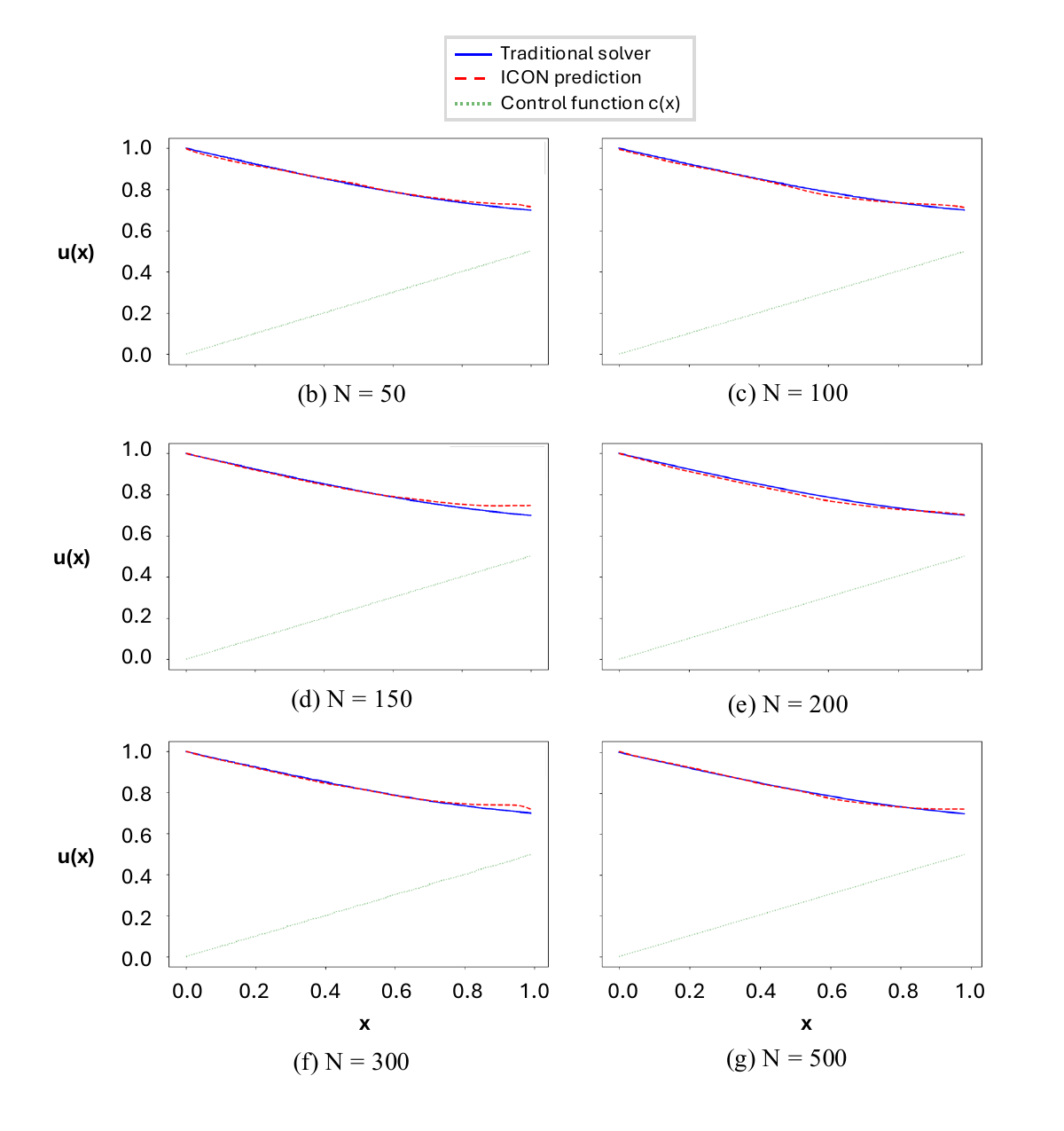}
    \end{subfigure}
\end{center}
\caption{Comparison of Poisson equation solutions for different domain sizes. (a) shows how inference time compares with domain length, while (b-g) show solutions for increasing domain lengths.}
\label{fig:poisson_comparison}
\end{figure}

A major weakness of the model is its scaling complexity. Though accuracy remains high, inference time increases as the length of the domain increases, corresponding to an increasing difference between ICON inference and a traditional solver, which remains constant (see Fig. \ref{fig:poisson_comparison}).

The linear scaling with domain length presents a trade-off in the model's practical applications. ICON is particularly vulnerable to coarse meshes, and this scaling suggests that for very large domains, the computational advantage of using neural methods may diminish.

However, this limitation should be viewed in context: For the domain sizes tested (up to N=500), the absolute inference times remain reasonable, with even the largest domain requiring only about 1.17 seconds. The high accuracy maintained across all domain sizes also suggests that the model's architectural design successfully preserves solution quality despite the increased computational load.
Future work could explore several avenues to address this scaling behavior:
\begin{itemize}
\item Investigating domain decomposition techniques that could allow parallel processing of different regions,
\item Developing hierarchical approaches that could handle different resolution levels efficiently, and
\item Exploring whether architectural modifications could reduce the dependency on domain length.
\end{itemize}

\section{Conclusion}

We demonstrate both the capabilities and limitations of In-Context Operator Networks for solving higher-order partial differential equations. While the model achieves performance comparable to previous benchmarks across 19 test problems and maintains high accuracy for in-distribution tasks, several important constraints emerge. Most notably, the linear scaling of inference time with domain length suggests potential limitations for very large-scale problems, though performance remains practical for domains tested up to \(N=500\).

Our analysis of out-of-distribution generalization, particularly through the heat equation case study, reveals that the model captures broad physical behaviors while struggling with specific features like boundary conditions and solution magnitude. This suggests that while ICONs can learn general principles of differential operators, their generalization capabilities may be more closely tied to pre-trained understanding than in-context examples.

Future work could address the identified scaling limitations through domain decomposition, hierarchical approaches, or architectural modifications. Additionally, improving out-of-distribution performance, particularly for boundary value problems and solution magnitude preservation, represents a promising direction for enhancing the model's generalization capabilities.
\pagebreak

\pagebreak
\begin{appendices}
\appendix
\section{Appendix}
\subsection{Model and training details}\label{appendix:training-details}
The model was trained on one node with 4 NVIDIA Quadro RTX 5000 GPUs on the Frontera cluster of the Texas Advanced Computing Center (TACC). Each GPU had 128GB of memory, and the model was trained with a batch size of 32. The training process utilized mixed precision to optimize memory usage and speed up computation. Information on TACC Frontera can be found at \url{https://tacc.utexas.edu/systems/frontera/}. The module runs a combination of JAX v0.6.0 and TensorFlow r2.16.1.
\begin{figure}[!htbp]
\begin{center}
\includegraphics[width=0.85\linewidth]{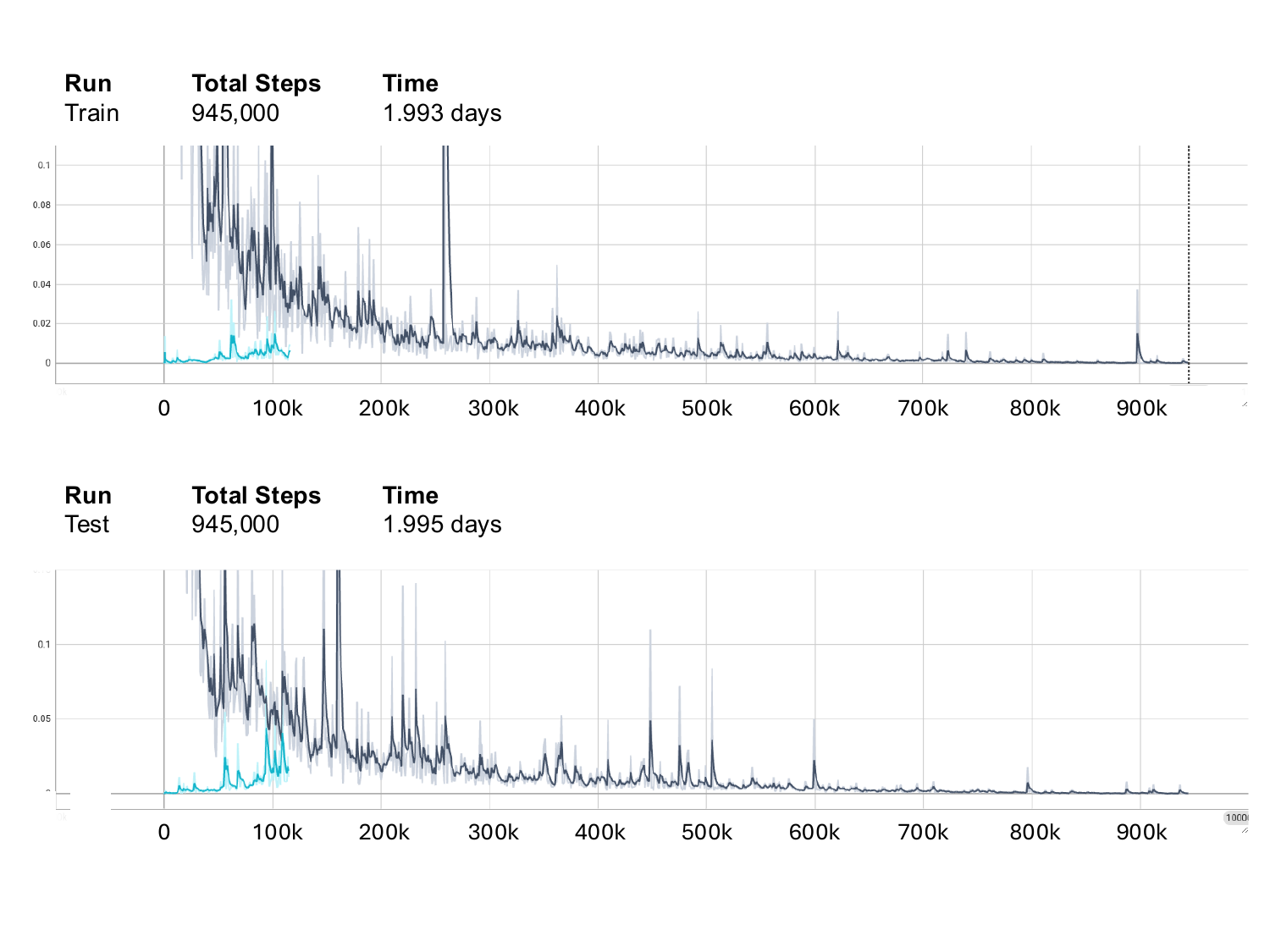}
\end{center}
\caption{Model loss for the \texttt{icon\_lm} model during training and initial inference. The upper curve shows the test loss, while the lower curve shows the training loss over the 50-hour, 1,000,000-step run (100 epochs, with 10,000 steps per epoch). The black curve represents the total loss during a continuous 90-epoch training run. The cyan curve is shown separately from the black curve to highlight the point at which training and test loss began to increase after 90 epochs.}
\label{fig:train-test}
\end{figure}
\begin{center}
\begin{tabular}{c|c}
\multicolumn{2}{c}{Transformer specifications} \\
\hline
    Layers & 6 \\
    Heads & 8 \\
    Head dimension & 256 \\
    Model dimension & 256 \\
    Dropout rate & 0 \\
    Widening factor & 4 \\
    & \(\implies\) hidden dimension = 1024\\
    Kernel initialization & Glorot uniform \\
    Attention function & vanilla
\end{tabular}
\end{center}

\begin{center}
\begin{landscape}
\subsection{ODE and PDE forms}\label{appendix:ode-pde-forms}

\begin{tabular}{|p{2cm}|p{5.0cm}|p{2cm}|p{2cm}|p{2cm}|p{5cm}|}
\hline
\textbf{Problem} & \textbf{Equation} & \textbf{Condition} & \textbf{QoI} & \textbf{Parameters} & \textbf{Operator} \\
\hline
ODE 1 & \( \frac{d}{dt}u(t)=a_1c(t)+a_2,\, u(0)=u_0 \) & \( c(t) \) & \( u(t) \) & \( a_1, a_2, u_0 \) & \( \mathcal{F}_{a_1,a_2,u_0}[c(t)] = u(t) \) \\
\hline
ODE 2 & \( \frac{d}{dt}u(t)=a_1c(t)u(t)+a_2,\, u(0)=u_0 \) & \( c(t) \) & \( u(t) \) & \( a_1, a_2, u_0 \) & \( \mathcal{F}_{a_1,a_2,u_0}[c(t)] = u(t) \) \\
\hline
ODE 3 & \( \frac{d}{dt}u(t)=a_1u(t)+a_2c(t)+a_3,\, u(0)=u_0 \) & \( c(t) \) & \( u(t) \) & \( a_1, a_2, a_3, u_0 \) & \( \mathcal{F}_{a_1,a_2,a_3,u_0}[c(t)] = u(t) \) \\
\hline
Damped oscillator & \( u(t)=A\sin\left(\tfrac{2\pi}{T}t+\eta\right)e^{-kt} \) & \( u(t)\big|_{t \in [0,0.5]} \) & \( u(t)\big|_{t \in [0.5,1]} \) & \( k \) & \( \mathcal{F}_{k}\!\left[u(t)\big|_{t \in [0,0.5]}\right] = u(t)\big|_{t \in [0.5,1]} \) \\
\hline
Poisson equation & \( u''(x)=c(x),\, u(0)=u_0,\, u(1)=u_1 \) & \( c(x) \) & \( u(x) \) & \( u_0, u_1 \) & \( \mathcal{F}_{u_0,u_1}[c(x)] = u(x) \) \\
\hline
Linear reaction-diffusion & \( -\lambda a u_{xx}(x)+k(x)u(x)=c, \)\newline \( u(0)=u_0,\, u(1)=u_1 \) & \( c(x) \) & \( u(x) \) & \( u_0, u_1, a, c \) & \( \mathcal{F}_{u_0,u_1,a,c}[k(x)] = u(x) \) \\
\hline
Nonlinear reaction-diffusion & \( -\lambda a u_{xx}(x)+ku^3(x)=c(x),\, \)\newline \( u(0)=u_0,\, u(1)=u_1 \) & \( c(x) \) & \( u(x) \) & \( u_0, u_1, k, a \) & \( \mathcal{F}_{u_0,u_1,k,a}[c(x)] = u(x) \) \\
\hline
Mean-field control & See \cite{icon_1_original} & \( \rho(t,x)\big|_{t=0} \) & \( \rho(t,x)\big|_{t=1} \) & \( g(x) \) & \( \mathcal{F}_{g(x)}\!\left[\rho(t,x)\big|_{t=0}\right] = \rho(t,x)\big|_{t=1} \) \\
\hline
Heat equation & \( u_t + (ku_x)_x + \alpha u = 0 \) & \( u(x,0) \) & \( u(x,\tau) \) & \( k, \alpha, \newline g_0(t), g_L(t) \) & \( \mathcal{F}_{k,\alpha,g_0,g_L}[u(x,0)] = u(x,\tau) \) \\
\hline
2-dim. 2nd-order linear PDE & \( au_{xx} + bu_{xy} + cu_{tt} + du_x + eu_y + fu = g(x,t) \) & \( g(x,t) \) & \( u(x,t) \) & \( a, b, c, \newline d, e, f, \newline f_0(x), f_1(x) \) & \( \mathcal{F}_{a,\dots,f,f_0,f_1}[g(x,t)] = u(x,t) \) \\
\hline
\end{tabular}
\end{landscape}

\end{center}

\end{appendices}


\begin{thebibliography}{99}

\bibitem{piml_2021}{\sc G. E. Karniadakis}, {\sc I. G. Kevrekidis}, {\sc L. Lu}, {\sc P. Perdikaris}, {\sc S. Wang}, and {\sc L. Yang}, {\em Physics-informed machine learning}, Nature Reviews Physics, 3 (2021), pp.~422--440.

\bibitem{weno}{\sc X.-D. Liu}, {\sc S. Osher}, and {\sc T. Chan}, {\em Weighted Essentially Non-oscillatory Schemes}, Journal of Computational Physics, 115 (1994), pp.~200--212.

\bibitem{deeponet_2021}{\sc L. Lu}, {\sc P. Jin}, {\sc G. Pang}, {\sc Z. Zhang}, and {\sc G. E. Karniadakis}, {\em Learning nonlinear operators via DeepONet based on the universal approximation theorem of operators}, Nature Machine Intelligence, 3 (2021), pp.~218--229.

\bibitem{icon_1_original}{\sc L. Yang}, {\sc S. Liu}, {\sc T. Meng}, and {\sc S. J. Osher}, {\em In-context operator learning with data prompts for differential equation problems}, Proceedings of the National Academy of Sciences, 120 (2023), pp.~e2310142120.

\bibitem{icon_2_finetune}{\sc L. Yang}, {\sc S. Liu}, and {\sc S. J. Osher}, {\em Fine-Tune Language Models as Multi-Modal Differential Equation Solvers}, arXiv preprint arXiv:2308.05061 (2024).

\bibitem{icon_3_pdegen}{\sc L. Yang} and {\sc S. J. Osher}, {\em PDE Generalization of In-Context Operator Networks: A Study on 1D Scalar Nonlinear Conservation Laws}, arXiv preprint arXiv:2401.07364 (2024).

\end{thebibliography}
\end{document}